\documentclass{article} 
\usepackage{iclr2025_conference, times}
\usepackage{graphicx}

\usepackage{amsmath,amsfonts,bm}

\def\eqref#1{equation~\ref{#1}}

\def\1{\bm{1}}

\DeclareMathAlphabet{\mathsfit}{\encodingdefault}{\sfdefault}{m}{sl}
\SetMathAlphabet{\mathsfit}{bold}{\encodingdefault}{\sfdefault}{bx}{n}

\usepackage{graphicx}
\usepackage{hyperref}
\usepackage{url}
\usepackage{booktabs}
\iclrfinalcopy
\title{Probing Mechanical Reasoning in Large Vision Language Models}

\author{%
  Haoran Sun\textsuperscript{1}, Qingying Gao\textsuperscript{1}, Haiyun Lyu\textsuperscript{2}, Dezhi Luo\textsuperscript{3,*}, Yijiang Li\textsuperscript{4,*}, Hokin Deng\textsuperscript{5}\thanks{Correspondence to Dezhi Luo (ihzedoul@umich.edu), Yijiang Li (yijiangli@ucsd.edu), Hokin Deng (hokind@andrew.cmu.edu)} 
  \\
  \textsuperscript{1}Johns Hopkins University, \textsuperscript{2}University of North Carolina Chapel Hill, \\ 
  \textsuperscript{3}University of Michigan, \textsuperscript{4}University of California San Diego, \textsuperscript{5}Carnegie Mellon University \\
All authors are affiliated with \href{https://growing-ai-like-a-child.github.io/}{GrowAI}
}

\begin{document}

\maketitle

\begin{abstract}
Mechanical reasoning is a hallmark of human intelligence, underpinning activities from everyday tool use to civil engineering. Endowing machines with this ability is thus a key step toward human-level AI. We evaluate 26 Vision-Language Models (VLMs) on 155 cognitive experiments spanning six domains—system stability, pulley systems, gear systems, the leverage principle, inertia and motion, and Liquid Mechanics. VLMs are showed the weakest on gear systems and Liquid Mechanics. Moreover, performance in these domains does not improve with model size, challenging simple scaling expectations. These results suggest that current attention-based architectures may lack mechanisms needed for mechanical reasoning, particularly those associated with mental simulation.

\end{abstract}

\section{Introduction}

Humans are uniquely capable of working with complex mechanical systems, from routine tasks such as assembling furniture to large-scale endeavors such as architectural design and advanced engineering \citep{harari2014sapiens}. These capabilities rest on the ability to reason about relations and interactions among physical objects—\emph{mechanical reasoning} \citep{clark2010supersizing, harman2011tool, vaesen2012cognitive}. While some animals exhibit limited forms of tool use or object interaction \citep{shumaker2011animal}, human mechanical reasoning is unparalleled in flexibility, sophistication, and creativity, enabling innovation, problem solving, and adaptation across environments \citep{allen2020rapid, allen2021learning}. Mechanical reasoning is therefore a cornerstone of human intelligence and a driver of technological and cultural progress. Endowing artificial systems with this ability is a vital step toward human-level performance in real-world settings. With rapid advances in large language models (LLMs) and their multimodal variants—\emph{Vision-Language Models} (VLMs)—mechanical reasoning provides a critical lens for assessing current capabilities and limitations.

Although mechanical reasoning is a high-level ability that emerges relatively late in development, it relies on more foundational strategies \citep{hegarty1988mentalmodels, kim1999perception, allen2021learning}. Decades of cognitive science research highlight the central role of \emph{mental simulation}—constructing and manipulating internal models of the world to guide inference \citep{hegarty2004mechanical}. Interviews and eye-tracking studies show that people dynamically build spatial representations of gears and pulleys to infer behavior while simulating motion \citep{hegarty1992mental, lehrer1998reasoning, kubricht2017intuitive, rozenblit2002mechanical}. By contrast, several studies suggest that the capacity to construct and exploit such visuospatial models remains underdeveloped in state-of-the-art LLMs/VLMs \citep{hao2023reasoning, mitchell2023debate, goddu2024llms, gao2024vision}. Because mental simulation requires model-based reasoning in the visuospatial domain, a thorough evaluation of VLMs’ mechanical reasoning can inform this debate. In this work, we leveraged the \textbf{MechBench} from the \textbf{CoreCognition} benchmark \citep{li2024core}, which comprises 155 single-image tasks adapted from cognitive psychology that spans six core domains of mechanical reasoning: System Stability, Pulley Systems, Gear Systems, Leverage Principle, Inertia and Motion, and Liquid Mechanics. Conducting the first systematic evaluation of 26 open- and closed-source VLMs against a human baseline, we reveal consistent human superiority, notable domain-specific performance gaps, and weak scaling-law effects in certain domains such as gear systems and Liquid Mechanics. Finally, we provide a cognitive-science-informed analysis linking VLM performance limitations to deficits in model-based reasoning, offering insights into architectural constraints and shortcut exploitation in current models.






\section{Methods}

\subsection{Experiment Design}

\begin{figure*}[t]
\centering
\includegraphics[width=1.0\textwidth]{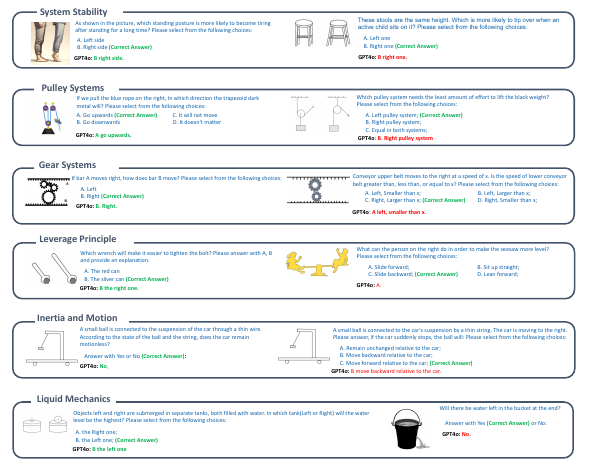}
\caption{\textbf{Sample tasks across the six MechBench domains.} For each domain, GPT\mbox{-}4o answered correctly on the left and failed on the right.}
\label{fig:fig1}
\end{figure*}

Mechanical reasoning has been widely explored in cognitive psychology and educational research. This study uses single-image adaptations of classic cognitive tasks from the literature to assess mechanical reasoning, focusing on six key domains: system stability, pulley systems, gear systems, leverage principle, inertia and motion, and liquid mechanics. Example cases on each domains are presented in Figure \ref{fig:fig1}. This categorization is intentionally designed to encompass the full range of mechanical reasoning, ensuring each category represents a distinct but fundamental physical understanding. Below, we provide explanations for each domain.

\paragraph{System Stability}
The understanding of stability is essential for reasoning about states of equilibrium within physical systems, such as predicting whether a stack of blocks will remain upright or collapse \citep{mccloskey1983intuitive}. In our experiments, models are presented with images of objects like stools with varying base widths or angles of inclination. The task involves selecting the most stable configuration. Stability reasoning encompasses factors such as the center of gravity, base area, and force distribution, making it a critical baseline for evaluating mechanical reasoning.

\paragraph{Pulley Systems}
Pulley systems are widely used in cognitive psychology and physics education to study how individuals reason about force and motion \citep{hegarty1994individual}, requiring an understanding of force distribution and machinery functions. For example, a simple task might involve determining which pulley system requires less effort to lift a weight. By testing VLMs’ ability to distinguish between fixed and movable pulleys and to predict object movement, we evaluate their capacity to infer about dynamic relationships within real-life scenarios.

\paragraph{Gear Systems}
Gear systems are deterministic mechanical setups governed by well-defined rules, such as adjacent gears rotating in opposite directions and gear ratios determining relative speeds. These properties make gears an ideal domain for testing logical and causal reasoning \citep{hegarty1988mentalmodels}. Tasks in this category involve analyzing diagrams of connected gears to predict their rotational direction and speed.

\paragraph{Leverage Principle}
The leverage principle illustrates the relationship between force, distance, and torque. Balance-scale experiments in cognitive psychology have shown how humans progressively develop an understanding of leverage through iterative learning and application \citep{peirce1992essential1}. Tasks in this category include determining how shifting weights on a seesaw or applying force to a lever affects balance.

\paragraph{Inertia and Motion}
Inertia and motion are dynamic aspects of mechanical reasoning that require understanding how forces influence the movement of objects over time. These concepts are central to Newtonian mechanics and intuitive physics \citep{mccloskey1983naive}. Human cognition integrates spatial and temporal information to make predictions about motion and forces, as seen in studies of tool use and physical reasoning \citep{allen2021learning}. Our experiments include scenarios such as predicting the trajectory of an object on a moving cart or identifying the kinetic energy distribution of a pendulum. These tasks thereby probe VLMs to integrate information about multiple aspects of the physical world.

\paragraph{Liquid Mechanics}
Liquid Mechanics involves understanding the behavior of liquids under various conditions, such as flow, external pressures, and volume changes. Although grounded in the intuitive understanding of fluid dynamics emerged very early in humans' cognitive development \citep{hespos2016nonsolid}, reasoning about liquid behaviors in mechanical systems require simultaneous consideration of geometry, force, and dynamics. These tasks represent a highly important domain of mechanical reasoning concerning specifically about liquid motions as opposed to solid objects alone.

\subsection{Examined Vision Language Models}
Recent advances in multi-modal learning have been driven by the unified modeling of visual and textual modalities using transformers \citep{li2019visualbert, xu2023bridgetower,tan2019lxmert, alayrac2022flamingo,radford2021learning}. With the rise of LLMs, state-of-the-art (SOTA) multi-modal LLMs (MLLMs) \citep{liu2024visual,li2023blip2} adopt open-source LLMs \citep{touvron2023llama, peng2023instruction,jiang2023mistral} and align visual features to the LLM embedding space \citep{li2023blip, fu2023mme, wu2024v, xu2024llava, shao2024visual, li2022more, li2025egoprivacy, brown2020language, achiam2023gpt, bai2023qwen, jaech2024openai, zhang2025unified, zhang2024pixels}. Progressively, MLLMs have demonstrated competitive performance in complex tasks involving high-level perception and reasoning \citep{li2024seed, team2023gemini, fu2023mme, openai2023gpt4}, such as spatial reasoning \citep{chen2024spatialvlm, cai2024spatialbot}, character recognition \citep{mori1999optical}, scene understanding \citep{cordts2016cityscapes, wang2023consistent, li2023diverse, chen2024sam, chen2024bridging}, action recognition \citep{jhuang2013towards, herath2017going} and prediction \citep{lan2014hierarchical, kong2022human}, reaching near-human performance.

We evaluated the mechanical reasoning abilities of three categories of VLMs. To ensure a fair comparison,  all VLMs are evaluated on their ability to reason over images and texts under a zero-shot generation task. A complete list of models is reported in the results section as shown in Figure \ref{fig:fig2}. Model size data are curated at the same time. The models are categorized as follows:

\begin{enumerate}
    \item \textbf{Open-source VLMs with Multi-Image Reasoning}:  
    Includes models with different sizes and other variants such as \texttt{CogVLM} Series\citep{hong2024cogvlm2}, \texttt{Qwen} series(Qwen-VL \citep{Qwen-VL}, Qwen-2 \citep{Qwen2VL}), and \texttt{Blip2} \citep{li2023blip2}, LLaVA-Next \citep{liu2024llavanext} , which are capable of reasoning over interleaved multiple images and texts.
    \item \textbf{Closed-source VLMs with Multi-Image Reasoning}:  
    Includes proprietary models such as GPT series \citep{gpt4o} ( \texttt{GPT-4v}, \texttt{GPT-4-turbo}, \texttt{GPT-4o-mini}), Gemini Series \citep{gemini}, and Claude Series \citep{claude}. These models also support reasoning across interleaved images and texts,
    \item     \textbf{Open-source VLMs with single-Image Reasoning}:  
   Includes models designed to process a single image alongside continuous text.  InstructBlip Series \citep{instructblip}, LLaVA Series \citep{liu2023improvedllava} \citep{liu2023llava} 
\end{enumerate}

In total, we aligned 26 models for evaluation. In order to analyze the reasoning abilities of VLMs, we asked the models to explain their answers after they have given the answers by adding "please provide an explanation" in the prompt.

\subsection{Human Baseline} 

We recruited a total of 9 participants, all of whom were college students proficient in English. Participants were instructed to skip any question that was ambiguously phrased or too complex to answer within 90 seconds. As a sanity check, a question was marked as failed if less than 80\% of the participants could answer within the time frame; otherwise, we modified the question, and new annotators completed the revised version. 


\section{Results}

\begin{figure*}[t]
\centering
\includegraphics[width=1.0\textwidth]{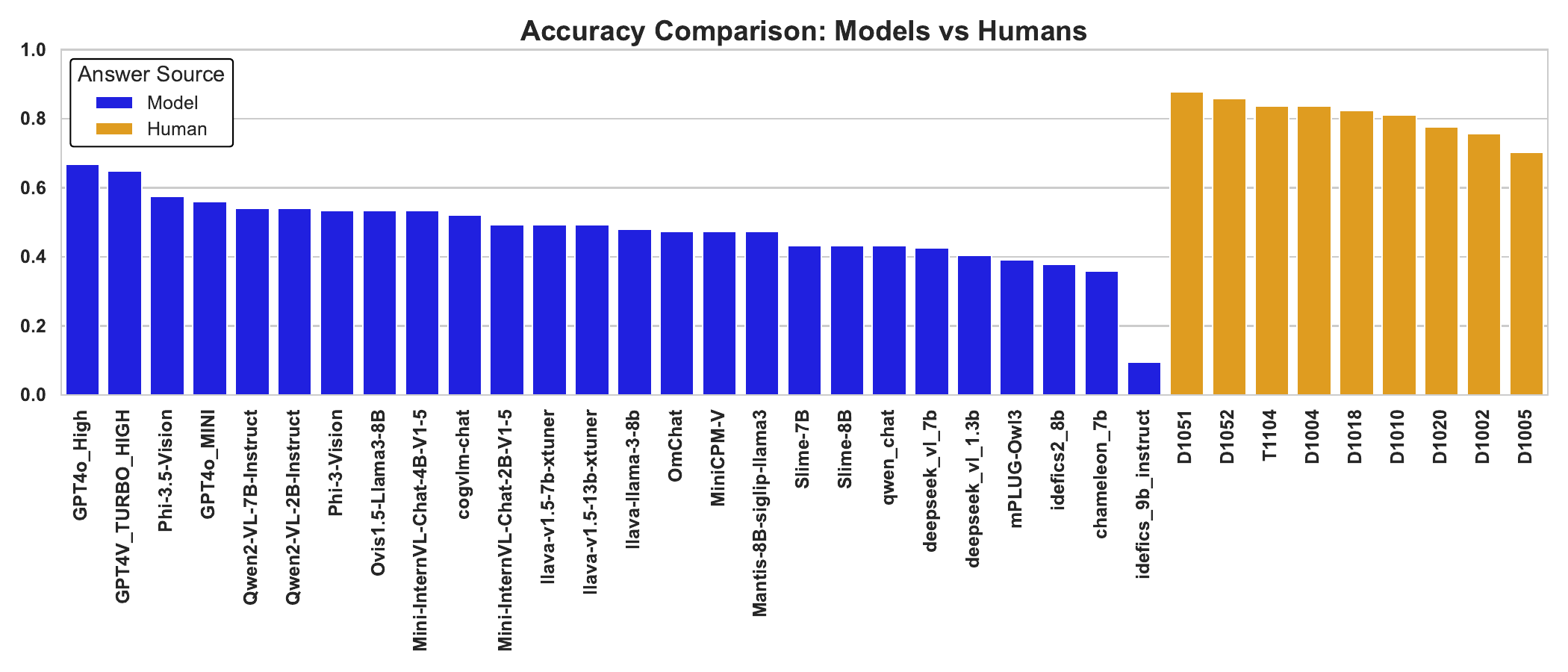}
\caption{\textbf{Model Performance on MechBench As Compared to Human Performance}}
\label{fig:fig2}
\end{figure*}

\begin{figure*}[t]
\centering
\includegraphics[width=0.9\textwidth]{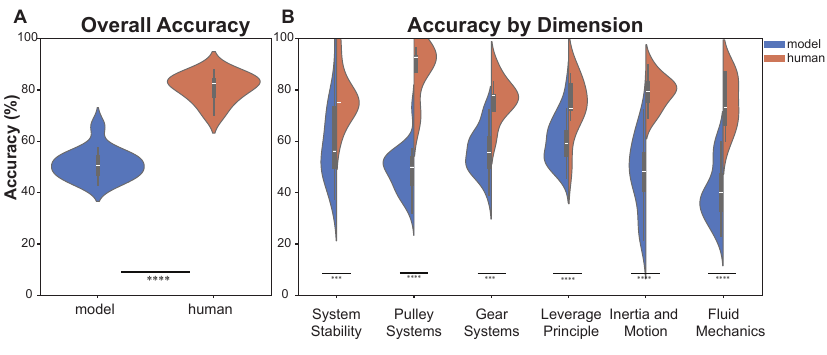}
\caption{\textbf{Overall and Domain-Wise Accuracy: Humans vs. Models}. A. For overall accuracy across tasks, human participants outperform models significantly (p \texttt{<} 0.0001). B. Human participants consistently outperform models in each domain (all categories p \texttt{<} 0.001).}
\label{fig:fig3}
\end{figure*}

Our study reveals a significant disparity between human and model performance across multiple evaluation dimensions. As shown in Figure \ref{fig:fig2} and \ref{fig:fig3}, humans consistently outperform models both in overall accuracy and in task-specific domains. These results highlight the limitations of current VLMs in replicating human-like reasoning in mechanical and intuitive physics tasks. Among the evaluating domains, Pulley Systems exhibits the largest performance gap, with human accuracy nearing 90\%, compared to model accuracy averaging around 50\%. 

\begin{figure*}[h]
\centering
\includegraphics[width=0.9\textwidth]{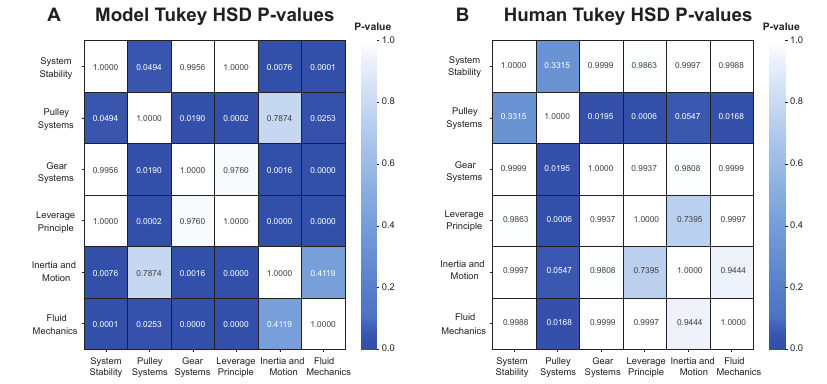}
\caption{\textbf{Comparison of Tukey HSD P-values for Model and Human Performance Across Different domains.}}
\label{fig:fig4}
\end{figure*}

\subsection{Human vs. Model Performance Across Each Domain}
The ANOVA test (\( F = 11.8111, p < 0.0001 \)) revealed significant differences in human performance across the six categories. Tukey HSD post-hoc analysis (visualized in the heatmap) shows that the performance of humans handling Pulley Systems is significantly different from the other five categories. The performance of humans handling the pulley system is significantly better than the other five categories. In contrast, ANOVA results for model performance (\( F = 1.6809, p = 0.1430 \)) indicated no significant differences across task categories. Tukey HSD analysis further reveals that models exhibited similar accuracy across all categories with the exception of Liquid Mechanics, which deviates significantly with other domains. Together, as shown in Figure \ref{fig:fig3}, the distribution of human performance is narrower, indicating more consistent accuracy, whereas the models display a wider range of variability. See Appendix~\ref{AppxDomain} for detailed analysis on representative model response and reasoning for each domain.


\subsection{Relationship Between Model Performance and Model Size}

\begin{figure*}[h]
\centering
\includegraphics[width=0.9\textwidth]{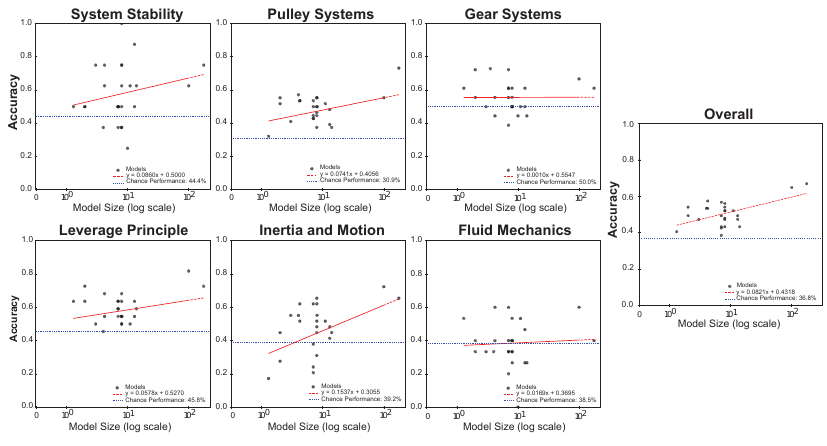}
\caption{\textbf{Relationship between model performance and model size.} The six panels on the left show how model size—measured by parameter count (log scale)—relates to accuracy across the six evaluation domains. The panel on the right reports overall accuracy across all tasks.}
\label{fig:fig5}
\end{figure*}

A widely held belief in the machine learning community is that an increase in a model's scale, measured by the number of parameters, leads to systematic improvements in its reasoning abilities \citep{sutton2019bitter, kaplan2020scaling}, a concept known as the scaling law. However, this assumption is an empirical observation without theoretical proof \citep{luo2025philosophical}. To evaluate whether the scaling law applies to mechanical reasoning, we further examined the relationship between model performance on mechanical reasoning tasks and model size, as measured by the number of parameters (Figure \ref{fig:fig5}). For overall accuracy across task, regression analysis yielded a formula $y = 0.0821x + 0.4318$ (p = 0.0053, R² = 0.2917). Models with parameter counts exceeding 10 billion, such as GPT-4o High and GPT4V-TURBO-HIGH,  distinctively outperformed smaller models. In most of the domains (System Stability, Pulley Systems, Leverage Principle, Inertia and Motion, Liquid Mechanics), accuracy increases with larger model sizes. However, Gear Systems and Liquid Mechanics show very weak scaling with slopes and R-squared values close to zero (Gear Systems: $y = 0.0010x + 0.5547$, p = 0.981628, R² = 0.000023; Liquid Mechanics: $y = 0.0169x + 0.3695$, p = 0.787137, R² = 0.003097). While model size often correlates with improved performance in many mechanical reasoning domains, some domains may not benefit significantly from increased model scale, indicating a potential distinction between the underlying mechanisms across different domains of mechanical reasoning.


\section{Discussions}


In this paper, we examine domain-specific patterns in VLM performance on mechanical reasoning tasks, showing that accuracy lags well behind human levels across all domains. While larger models tend to achieve higher overall accuracy, the strength of this relationship varies markedly by domain. Notably, VLMs' performance on Gear Systems and Liquid Mechanics remains close to chance and does not improve with increased model size, serving as a counterexample to the scaling law hypothesis in machine learning \citep{sutton2019bitter, kaplan2020scaling}. Such a trend indicate that the underlying architecture of these models is yet to be able to support certain mechanisms that are required for a system to have mechanical reasoning abilities regarding Gear System and Liquid Mechanics. Similar trends have previously been reported in VLMs' performance on level-2 perspective-taking tasks, in which the models are asked to tell how another agent would see the same spatial arrangement in a perspective different from their own \citep{gao2024vision, piaget1969psychology, Moll2011look}. Interestingly, like mechanical reasoning, such ability is also known to require mental simulation in humans, which is the internal construction and manipulation of dynamic models of the world to predict or explain outcomes \citep{hegarty2004mechanical, zhao2016nine}. In cognitive science, mental simulation is often considered a concrete instantiation of model-based reasoning, where inferences are produced by running an internal model to project likely outcomes, rather than relying solely on pre-learned patterns or correlations \citep{johnson1983mental, lecun2022path, balaban2025physics}. Together, these findings suggest that VLMs may lack the capacity for such simulation-driven, model-based reasoning, possibly highlighting a fundamental limitation in the architecture of current foundational models \citep{mitchell2023debate,goddu2024llms,li2024core}.


There is, however, an important concern with this interpretation: if VLMs’ limitations in mechanical reasoning stem from an inability to perform model-based reasoning, why do they not also underperform on domains like Pulley Systems, which likewise rely on such reasoning? Two explanations may account for this discrepancy. First, the role of mental simulation is mediated by lower-level cognitive abilities often termed intuitive physics—expectations about object relations and motion \citep{mccloskey1983intuitive, kubricht2017intuitive}. Intuitive physics draws on, but is not reducible to, model-based reasoning; it depends on the domain-specific application of physical primitives such as motion, gravity, and states of matter \citep{kaiser1986intuitive, kubricht2017intuitive}. Different domains recruit different primitives. For example, Liquid Mechanics relies on priors about liquids that differ from those governing solids \citep{kawabe2015seeing}. Models may therefore exhibit some model-based reasoning yet fail to apply the appropriate primitives in certain domains, yielding a gap in performance.

Second, models may exploit shortcuts that bypass genuine simulation. Pulley systems are heavily represented in educational materials and typically involve few components (e.g., fixed and movable pulleys) with limited combinatorics, enabling spurious correlations between the benchmarked domain and patterns present in training data \citep{bender2021dangers, bleeker2024demonstrating}. Apparent competence may thus reflect pattern matching rather than step-by-step reasoning. Consistent with this account, very large models (e.g., GPT-4o) perform particularly well on pulley tasks, whereas smaller models lag.

\section{Conclusion}

In this paper, we presented the first comprehensive assessment of VLMs' mechanical reasoning abilities. Drawing on experimental paradigms from cognitive science literature, our findings reveal that current VLMs fail to fully acquire the necessary competencies for mechanical reasoning across various domains. Furthermore, the observation that performance on certain tasks does not improve with increased model size underscores potential fundamental limitations in existing model architectures. In line with previous research, we specifically highlighted model-based reasoning—the ability to mentally simulate dynamic relations within physical systems. However, given that mechanical reasoning is a high-level cognitive ability reliant on a range of basic cognitive processes \citep{kaiser1986intuitive,hegarty1988mentalmodels,allen2021learning}, the precise mechanisms underlying these limitations remain unclear. Further exploration of these mechanisms could provide crucial insights into the constraints of current AI models and inform the development of more capable architectures for mechanical reasoning tasks.

\newpage
\bibliography{iclr2025_conference}
\bibliographystyle{iclr2025_conference}

\vspace{25pt}

\newpage

\appendix{\large\bf{Appendix}}

\section{Detailed Analysis of Model Performance on Each Domain of Mechanical Reasoning}
\label{AppxDomain}

\subsection{System Stability}

\begin{figure}[h]
  \centering
  \includegraphics{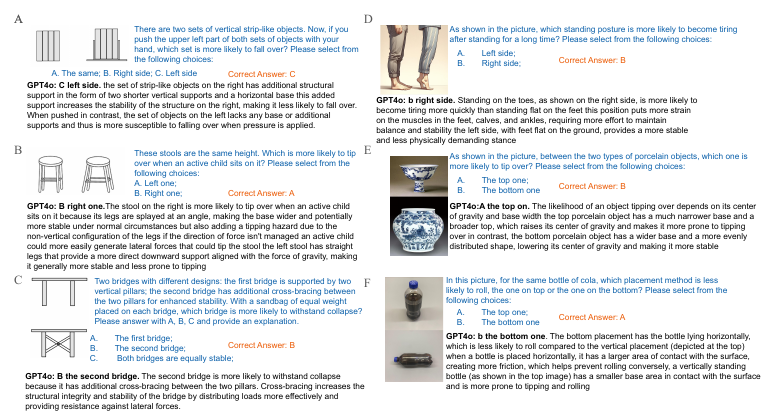}
  \caption{VLMs performance on reasoning about mechanical system stability in MechBench}.
  \label{fig:fig6}
\end{figure}

Models excel at identifying objects in images, and they not only understand what the objects are, the models also successfully recognize their mechanical states as well. Also, the models could effectively connect mechanical descriptions with the corresponding mechanical scenarios (Figure \ref{fig:fig6}). In Experiments \ref{fig:fig6}B and \ref{fig:fig6}F, the models could recognize the two chairs and the two bottles, and their mechanical situations; however, the models still failed to provide the correct answers in the experiments. For instance, in Experiment \ref{fig:fig6}B, the model explains, "the stool on the right is more likely to tip over when an active child sits on it because its legs are splayed at an angle, making the base wider and potentially more stable under normal circumstances but also adding a tipping hazard due to the non-vertical configuration of the legs". The model correctly notices that it's the angle of the leg that matters for the stability of the system. However, it reasons completely the opposite way to correct answers. When the leg's angle wider, it's actually more stable. It's an intuitive physical rule problem for humans but the models fail, even though they still demonstrate step-by-step reasoning abilities in this case. In Experiment \ref{fig:fig6}F, the model is correct that the bottle on the bottom "has a larger area of contact with the surface, creating more friction, which helps prevent rolling". However, the model fails to realize one bottle is standing, and one bottle is rolling, and sliding friction and rolling friction are completely different.

\newpage

\subsection{Pulley Systems}

\begin{figure}[t]
  \centering
  \includegraphics{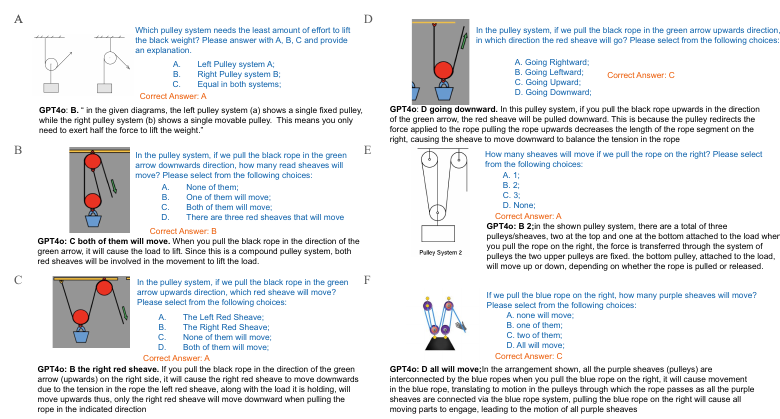}
  \caption{VLMs performance on reasoning about pulley systems in MechBench}.
  \label{fig:fig7}
\end{figure}

We find that current VLMs struggle to handle pulley systems (Figure \ref{fig:fig7}).  We observe that there are generally three failures in VLMs reasoning about pulley systems: first, VLMs are not able to identify which are the movable pulleys in the system, and second, VLMs exhibit relatively low accuracy in determining whether an object is rising or falling through pulley systems.

VLMs perform poorly in recognizing movable pulley systems. In Experiment \ref{fig:fig7}A, the image includes a standard single movable pulley system and a standard single fixed pulley system. The question "Which system requires less effort?" is essentially asking whether the model can correctly select the movable pulley. Clearly, the model failed in its selection, as it straightforwardly provided an incorrect answer in its explanation. VLMs also struggle in predicting whether a suspended weight is being lifted or lowered through a pulley system. Experiments \ref{fig:fig7}B-F either directly or indirectly reflect this issue, with Experiment \ref{fig:fig7}D being the most direct and concise. In Experiment D, the weight is directly attached to the movable pulley, and by pulling the other end of the rope, the pulley and the weight are lifted. However, the model's response was the exact opposite of the correct answer. In its explanation, the model seemed to imply that the pulley was not fixed (though it did not explicitly state that it was a movable pulley), and the physics it provided was entirely incorrect. Therefore, we can hypothesize that the model's poor performance in predicting the weight's movement may be due to its limited ability to recognize movable pulleys. However, the specific reasons require further experiments to be analyzed in detail.

The above two issues confirm that VLMs still have limitations in recognizing pulley systems, whereas for individuals with some mechanical experience, identifying simple pulley systems through basic diagrams is not difficult (especially in the case of Experiments \ref{fig:fig7}A and \ref{fig:fig7}D).

\newpage

\subsection{Gear Systems}

\begin{figure}[t]
  \centering
  \includegraphics{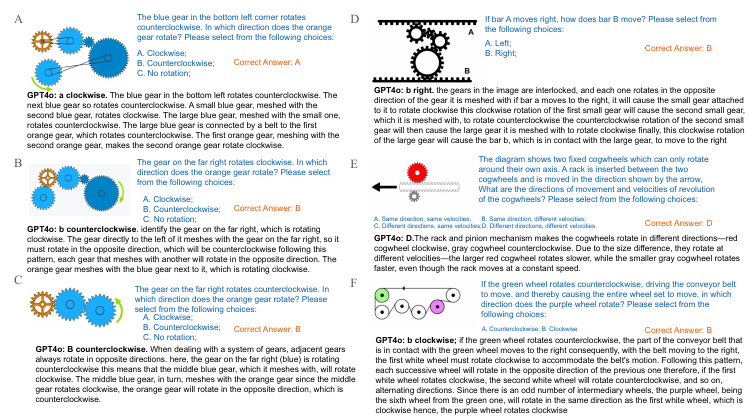}
  \caption{VLMs performance on reasoning about gear systems in MechBench}.
  \label{fig:fig8}
\end{figure}

On gear and conveyor belt problems, VLMs are able to give highly accurate judgments (Figure \ref{fig:fig8}). In particular, VLMs can reliably determine the rotation of one pulley based on the direction of another. 

Compared to pulley systems, we speculate several reasons why VLMs perform better in gear systems. First, gear problems typically involve simple mechanical setups with fixed scenarios, meaning the layout and interactions between gears are often straightforward. This reduces the complexity for the model when analyzing the system, as it does not need to account for too many variables or unpredictable changes in the structure. In other words, a gear system could be reduced into a logical diagram, and solved correctly using that diagram. 

Secondly, gear problems rely on a few key physical principles. Specifically, there are two important rules: adjacent gears always rotate in opposite directions, and smaller gears rotate faster than larger ones. Unlike problems that involve complex mathematical calculations, these rules are straightforward for the model to understand and apply consistently. As a result, the model can accurately analyze and predict gear behavior. Similarly, these physical rules can be reduced to logical rules, allowing the model to solve the problem using fixed logical principles.

\newpage

\subsection{Seesaw-like systems and Leverage Principle}

\begin{figure}[t]
  \centering
  \includegraphics{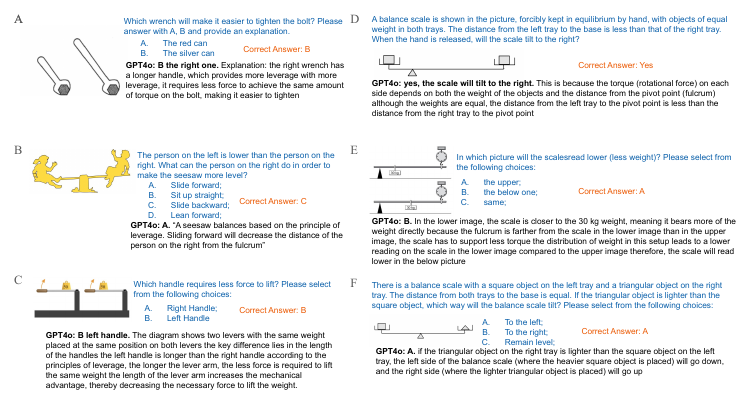}
  \caption{VLMs performance on reasoning about seesaw-like systems and leverage principle in MechBench}.
  \label{fig:fig9}
\end{figure}

We observe diverse behaviors in VLMs on solving seesaw-like mechanical systems and applying leverage principle (Figure \ref{fig:fig9}). Experiments \ref{fig:fig9}A, \ref{fig:fig9}C, \ref{fig:fig9}D, and \ref{fig:fig9}F involve relatively simple scene-based question answering and basic extrapolation. Experiments \ref{fig:fig9}A and \ref{fig:fig9}D are straightforward question-answer tasks. As long as the model identifies the key information in the images—such as the length of the wrench in Experiment \ref{fig:fig9}A or the unequal torques in Experiment \ref{fig:fig9}D—it can easily solve the problems. Experiments \ref{fig:fig9}C and \ref{fig:fig9}F involve simple extrapolations based on scene information, like understanding how changes in the length of the resistance arm affect force in Experiment \ref{fig:fig9}C or predicting the outcome when the weights on a balance scale are unequal in Experiment \ref{fig:fig9}F. It is evident that the model performed exceptionally well in these experiments, providing both correct answers and detailed explanations. However, in Experiments \ref{fig:fig9}B and \ref{fig:fig9}E, VLMs failed to give the correct answers. We suspect that VLMs struggle when the mechanical reasoning processes, particularly applying the leverage principle and solving seesaw-like systems, require multiple steps. For example, in Experiment \ref{fig:fig9}B, while VLMs recognize it is a seesaw-like system, they fail to understand how to operate within this system: sliding backward would give more leverage. This reasoning process may require backtracking through several steps: first, identifying the seesaw-like system; second, understanding how to apply leverage; third, realizing that sliding backward increases leverage; fourth, recognizing that more leverage is needed to level the system; and finally, concluding that sliding backward is necessary. The same issue arises in Experiment \ref{fig:fig5}E. 

\newpage

\subsection{Inertia and Motion}

\begin{figure}[t]
  \centering
  \includegraphics{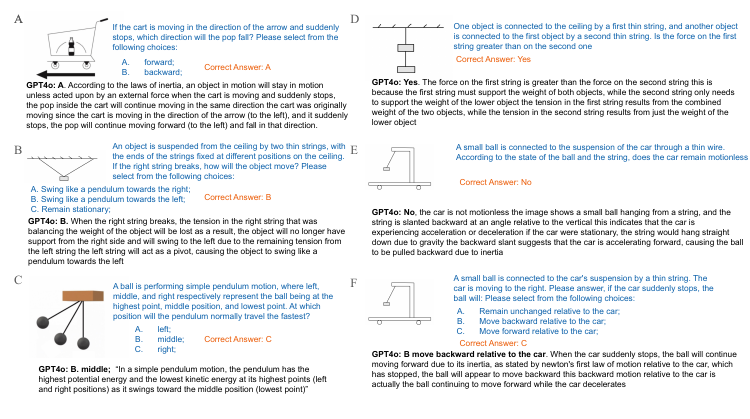}
  \caption{VLMs performance on reasoning about inertia and motion in MechBench}.
  \label{fig:fig10}
\end{figure}

We also observe very diverse behaviors in VLMs' understanding of inertia and motion (Figure \ref{fig:fig10}). Similar to seesaw-like systems, VLMs are able to identify the mechanical situations in the problem settings. However, they lack the ability to effectively predict the next step based on the current scene.

Experiment \ref{fig:fig10}E involves scene-based question. The model not only identified that the cart was not stationary, but it also further analyzed that if the object suspended on the cart were stationary, it would fall vertically due to gravity. However, the depicted scene likely indicates that the cart is accelerating forward. Experiments \ref{fig:fig10}A, \ref{fig:fig10}B, and \ref{fig:fig10}D, on the other hand, are cases of simple extrapolation. They are categorized as "simple extrapolation" because the scenarios are straightforward and involve a single, clear change. For example, in Experiment \ref{fig:fig10}A, the change involves the cart suddenly stopping—a relatively simple scene (involving only two objects, the drink and the cart, with straight-line motion and vertical force equilibrium). Similar patterns apply to Experiments \ref{fig:fig10}B and \ref{fig:fig10}D.

Experiments \ref{fig:fig10}C and \ref{fig:fig10}F, however, involve more complex reasoning related to inertia and motion prediction. Experiment \ref{fig:fig10}F is a derivative question from Experiment \ref{fig:fig10}E, asking what would happen if the cart suddenly stopped. The suspended object would continue moving forward due to inertia. Although the scene in Experiment \ref{fig:fig10}F is similar to Experiment \ref{fig:fig10}E, and the question is similar to that in Experiment \ref{fig:fig10}A, it involves more objects (the pulling rope, the cart, and the pulled object), more physical principles, and a sudden change in force (as the rope loses tension). The model clearly struggled with this problem, providing an incorrect answer and an explanation that did not meet expectations. Notably, Experiment \ref{fig:fig10}C asks a common-sense question based on pendulum motion. The image in Experiment \ref{fig:fig10}C marks three points in the half-arc of the pendulum's swing: the highest point, the quarter-arc point, and the lowest point. This information is also specified in the prompt. However, the model gave an incorrect answer, and its explanation was confused. The model completely misunderstood the designated points in the diagram and in the prompt, mistaking the first and third points as the left and right highest points of the pendulum's motion and the second point as the lowest point. One possible reason for this error is that the model may have mistaken the half-arc pendulum motion in Experiment \ref{fig:fig10}C for a full pendulum motion. A deeper explanation might suggest that the model has issues with perceptual constancy. The image in Experiment \ref{fig:fig10}C might have been misinterpreted as a 3D scene, and this visual misperception could be the root of the error. This hypothesis requires further experimentation to be confirmed.

\newpage

\begin{figure}[t]
  \centering
  \includegraphics{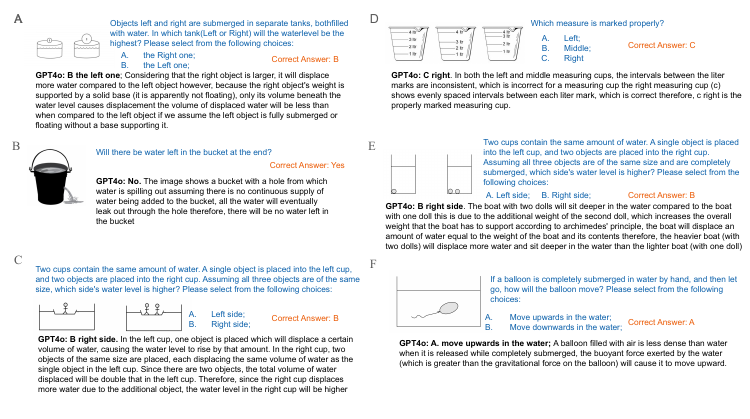}
  \caption{VLMs performance on reasoning about Liquid Mechanics in MechBench}.
  \label{fig:fig11}
\end{figure}

\subsection{Liquid Mechanics}

The fluid-related experiments involved properties such as fluid flow, buoyancy, and volume. In fluid-related systems, VLMs still face the aforementioned issues, particularly the challenge of complex inference. However, a notable highlight is that VLMs have demonstrated impressive scene understanding and detail-capturing abilities.

Experiments \ref{fig:fig11}B and \ref{fig:fig11}C highlighted the model's weaker inference abilities. In Experiment \ref{fig:fig11}B, the hole in the bucket was positioned at the middle of the bucket's wall. Clearly, once the water level drops to the level of the hole, water will stop flowing out, meaning some water will remain in the bucket. However, the model failed to capture this crucial detail about the hole's location, leading to an incorrect answer. Similarly, in Experiment \ref{fig:fig11}C, although the model provided the correct answer, its explanation was incorrect.

It is worth noting that the model excelled in image comprehension and detail recognition, especially in Experiments \ref{fig:fig11}D and \ref{fig:fig11}F. In Experiment \ref{fig:fig11}D, the measuring cup had a narrow base and a wider top, meaning the scale markings could not be evenly spaced. The model successfully captured these details, including the design of the measuring cup, the distribution of the scale, and the corresponding numerical values and units, offering a detailed explanation. The same can be said for Experiment \ref{fig:fig11}F. The model accurately identified that the balloon was fully inflated with gas and incorporated the concepts of gas and liquid density to explain the current state and predict the next step of the experiment.




\end{document}